\pdfoutput=1

\documentclass[11pt]{article}

\usepackage{acl}

\usepackage{times}
\usepackage{latexsym}

\usepackage[T1]{fontenc}

\usepackage[utf8]{inputenc}

\usepackage{microtype}

\usepackage{enumitem}
\usepackage{lipsum}
\usepackage{amsmath}
\usepackage{booktabs}
\usepackage{bm} 
\usepackage{nicefrac}
\usepackage{graphicx}
\usepackage{multicol}
\usepackage{listings}
\usepackage{xcolor}
\usepackage{tcolorbox}
\usepackage{bm}
\usepackage{inconsolata}

\lstdefinestyle{mystyle}{
    basicstyle=\small\ttfamily, 
    frame=single,
    backgroundcolor=\color{gray!10},
    rulecolor=\color{black},
    breaklines=true,
    showstringspaces=false
}

\usepackage[main=british]{babel}

\useshorthands*{"}
\defineshorthand{"-}{\babelhyphen{hard}}
\defineshorthand{"=}{\babelhyphen{–}}
\defineshorthand{"/}{\babelhyphen{/}}

\newcommand*{\GPTIV}{\texttt{GPT-4}}
\newcommand*{\ROUGEL}{\textsc{Rouge}"-L}
\newcommand*{\BERTSCORE}{\textsc{BertScore}}
\newcommand*{\PA}{$\text{PA}$}

\newenvironment{finding}[1]{\trivlist\item\relax \textbf{Finding~#1:} \itshape}{\endtrivlist\noindent}

\title{How Reliable Are Automatic Evaluation Methods\\ for Instruction-Tuned LLMs?}

\author{
    Ehsan Doostmohammadi\textsuperscript{$\dagger$} \quad 
    Oskar Holmström\textsuperscript{$\dagger$}  \quad 
    Marco Kuhlmann \\
    Linköping University \\
    \normalsize \texttt{ehsan.doostmohammadi@liu.se}
}

\begin{document}
\maketitle
\def\thefootnote{$\dagger$}\footnotetext{Equal contribution}\def\thefootnote{\arabic{footnote}}
\begin{abstract}
Work on instruction"-tuned Large Language Models (LLMs) has used automatic methods based on text overlap and LLM judgments as cost"-effective alternatives to human evaluation.
In this paper, we perform a meta"-evaluation of such methods and assess their reliability across a broad range of tasks.
In evaluating how well automatic methods align with human evaluations, correlation metrics are the most commonly employed method despite their inherent limitations when dealing with ties and different scales.
To address these shortcomings, we use Pairwise Accuracy as an alternative to standard correlation measures.
We observe that while automatic evaluation methods can approximate human ratings under specific conditions, their validity is highly context"-dependent.
Specifically, the simple \ROUGEL\ metric correlates very well with human ratings for short"-answer English tasks but is unreliable in free"-form generation tasks and cross"-lingual scenarios.
The effectiveness of the more advanced method of using \GPTIV\ as a judge diminishes significantly if reference answers are not included in the prompt, which is the scenario where this method has the potential to provide the most value compared to other metrics.
Our findings enhance the understanding of how automatic methods should be applied and interpreted when developing and evaluating instruction"-tuned LLMs.
\end{abstract}

\section{Introduction}

A key strength of the current generation of Large Language Models (LLMs) is their capacity to learn new tasks from instructions, either in"-context \cite{mishra-cross,sanh2022multitask,wei2022finetuned} or in a dedicated fine"-tuning phase \cite{wang2022super}.
The field has also seen the development of methods to adapt LLMs to new languages, for example, through continued fine"-tuning \cite{muennighoff-etal-2023-crosslingual}, alignment with translation pairs \cite{ranaldi2023empowering}, and instruction tuning on additional languages \cite{chen2023monolingual, kew2023turning}.

The gold standard for evaluating generative tasks is human annotation, but this scales poorly due to high costs and time constraints.
Consequently, the most common approach for assessing generative LLMs is using automated evaluation techniques.
Among these, two popular methods are measuring text overlap with \ROUGEL\ \cite{lin-2004-rouge} and utilizing existing LLMs as automatic judges \cite{zheng2023judging}; however, these methods only approximate human judgment, prompting questions about their reliability.
While previous research has found that automatic evaluation methods correlate well with human assessments \cite{wang2022super,zheng2023judging}, it is important to recognize that these findings generalize over tasks of very different types and in different languages.
Additionally, correlation measures may not provide reliable estimates of alignment with human ratings, as they are limited in their ability to deal with ties and constant scores, which are common in human annotations \citep{deutsch-etal-2023-ties}.

In this paper, we provide a thorough analysis of two widely"-used automatic methods for approximating human judgments, \ROUGEL\ and LLM"-as"-a"-judge.
Additionally, we experiment with \BERTSCORE, a semantic text similarity measure, to assess its potential utility.
We study the reliability of the three measures across a broad range of English"-language tasks.
We also perform experiments on Swedish as an initial study on the reliability of these metrics across languages.
Instead of using correlation measures, we employ Pairwise Accuracy \citep{deutsch-etal-2023-ties} to quantify the alignment with human ratings.
Our overall goal is to increase our understanding of the reliability of automatic evaluation methods and to establish guidelines regarding their appropriate usage.

\noindent
Our contributions can be summarized as follows:

\begin{itemize}[leftmargin=1em]

\item We adopt Pairwise Accuracy (PA) with tie calibration \cite{deutsch-etal-2023-ties} to enable robust comparisons between metrics, as we observe a high prevalence of tied ratings which renders common metrics, such as Kendall's $\tau$ and Spearman's $\rho$, unreliable.

\item We show that \GPTIV\ aligns well with human judgments when gold reference answers are available. However, its reliability diminishes in the absence of these references, where it shows an overly positive bias. This is especially problematic for free"-form tasks, since \GPTIV\ is commonly used in such settings.

\item We find that \GPTIV, while being the best tool for evaluating generations, can be replaced by faster and far less costly alternatives under certain conditions. In particular, we show that \ROUGEL\ offers a cost"-effective alternative to \GPTIV\ for short"-answer tasks, while \BERTSCORE\ shows promising results in long"-answer tasks.

\item We observe a decrease in alignment with humans in non"-English tasks for \ROUGEL\ and \GPTIV\ in situations where it does not have access to gold references. This suggests that it could be challenging to use automatic evaluation methods for lesser"-resourced languages.

\end{itemize}

\section{Related Work}

We start by reviewing the research on the automatic evaluation of generated text, the use of LLMs as evaluators, and the methods applied to assess the alignment of metrics with human preferences.

\subsection{Automatic Evaluation}

For short"-form tasks such as multiple"-choice question answering, assessing the quality of model outputs appears feasible through standard classification metrics like accuracy and F\textsubscript{1}-score \cite{li2023bactrian,lai2023chatgpt}.
While such an evaluation can be precise, it is rather strict and can only provide a fair performance assessment if the model does not deviate from the instructed format.
However, this easily happens as the tasks diverge from the training data or get more complex.
Surface"-level similarity measures such as \ROUGEL\ \cite{honovich2022unnatural,wang2022self,mishra-cross,yin2023did,lai2023chatgpt,li2023m} are more forgiving regarding formatting inconsistencies, but still lack the sophistication to be effective in tasks where free"-from answers are expected.

\subsection{Evaluation Using LLMs}

An increasingly common method for evaluating instruction fine"-tuned models is to use powerful LLMs as automatic judges \cite{peng2023instruction,doi:10.1073/pnas.2305016120,chen2023monolingual,kew2023turning}.
\citet{zheng2023judging} propose three different variations: (1) pairwise comparison, which asks the LLM to choose its preferred answer or declare a tie; (2) single"-answer grading, in which the LLM is asked to assign a score to an answer; and (3) reference"-guided grading, in which the model is provided with a reference solution (if available).
An approach similar to the second one is used by M\textsuperscript{3}IT \cite{li2023m} to evaluate the accuracy, relevance, and naturalness using \GPTIV\ in a multimodal scenario.

\subsection{Meta-Evaluation}

Human evaluation is the gold standard of assessment in natural language processing, but is not widely used in the literature due to its high costs.
Instead, authors have turned to automatic evaluation measures that correlate well with human judgments.
\citet{wang2022super} find a consistently strong correlation between \ROUGEL\ scores with accuracy across different models and task types, indicating that it is a good proxy for accuracy in classification tasks with short outputs.
For machine translation, 
\citet{zhang2020bertscore} show that \BERTSCORE\ is better correlated to human judgments than previous metrics, but \citet{hanna-bojar-2021-fine} also identify setups where it fails.

The recent work on automating evaluation processes and leveraging LLMs has demonstrated substantial agreement with human ratings. 
\citet{zheng2023judging} show that \GPTIV's judgments align with human evaluations at over 80\% agreement, reaching levels comparable to human"=human agreement.
\citet{zhou2023lima} also report agreement levels between \GPTIV\ and human annotators on a par with human"=human agreements.
There is also work on the meta"-evaluation of automatic metrics for chat and summarization \cite{shen-etal-2023-large,chiang-lee-2023-closer} using different criteria, and on aligning language model evaluations better with human preferences, such as \citet{liu-etal-2024-calibrating-llm} and \citet{chan2024chateval}.



\section{Methodology}

In this section, we present the data and instruction-tuned models used in our study.
We provide an overview of the automatic metrics we employed to assess model performance and detail our approach to conducting a meta"-evaluation of these metrics.

\subsection{Data}

As our training data, we use the Cleaned Alpaca Dataset\footnote{\href{https://github.com/gururise/AlpacaDataCleaned}{https://github.com/gururise/AlpacaDataCleaned}}, which corrects errors found in the original Alpaca \cite{alpaca}.
For testing, we use Natural Instructions v2 (NIv2) \cite{mishra-cross, wang2022super}, which spans a diverse range of tasks, including classification, question answering, free"-form text generation, and reasoning.
This enables fine"-grained testing.

\paragraph{Sample Selection}

Because of our limited annotation budget (cf.\ \S\ref{sec:EvaluationMethods}), we select 20 from the 119 (English"-language) tasks available in NIv2.
We aim to find tasks that (a)~cover a range of difficulty levels, (b)~involve both short and long free"-form answers, and (c)~are diverse in task types while leaving some type overlap for control purposes.
For a full description of the selected tasks, we refer to Appendix~\ref{app:tasks}.
From each task, we pick 15~random samples, leaving us with 300 samples in total.

\paragraph{Translation}

To study the metrics' reliability across the language dimension, we translate both our training and our test data to Swedish using \texttt{GPT-3.5-turbo}.
The prompt template and hyperparameters used for translation can be found in Appendix~\ref{app:prompts}.
Previous work has shown that the automatic translation of the Alpaca dataset produces high"-quality results with low noise levels \cite{holmstrom-doostmohammadi-2023-making,li2023bactrian}.
In addition to the two monolingual train datasets, we create an equally"-sized English"=Swedish bilingual train set by replacing a random 50\% of the samples in the Cleaned Alpaca Dataset with their Swedish translations.
Our purpose with this bilingual training set is to conduct a controlled study with more diverse bilingual data.

\subsection{Instruction Tuning}
\label{sec:method}

We instruction"-tune three base models in this study: \texttt{LLaMA2-7b}, \texttt{LLaMA2-13b} and \texttt{GPT-SW3-6.7b}.
Our selection accounts for different model sizes, pretraining languages, and performance.
\begin{itemize}[leftmargin=*]
\item \texttt{LLaMa2} \cite{touvron2023llama} is trained on mainly English data; only 0.15\% of the pretraining data is Swedish.
We use both the 7B and the 13B parameter versions of the model.
\item \texttt{GPT-SW3} \cite{ekgren2023gpt} is a \texttt{GPT-2}"-based language model mainly trained on North Germanic languages and English, where 26\% is Swedish, and 40\% is English.
\texttt{GPT-SW3} exhibits the lowest perplexity on Swedish (see Appendix~\ref{app:ppl}),
which is unsurprising as \texttt{LLaMA2} models have seen vastly fewer Swedish tokens during pretraining.
\end{itemize}

\noindent
For instruction tuning, we use the same training settings, hyperparameters, and prompts as \href{https://github.com/tatsu-lab/stanford}{Alpaca}, and use DeepSpeed \cite{10.1145/3394486.3406703} with the same configuration for all models.
For more details regarding the implementation, see Appendix~\ref{app:imp}.


\paragraph{Naming Scheme}

We instruction"-tune each of our three base models on the three training datasets (English, Swedish, and bilingual) and test it on either the single training language (for monolingual models) or both languages (for bilingual models).
This gives us a combined total of 12~different configurations for our experiments.
Throughout the paper, we refer to these by the name of the base model, model size, training dataset, and testing language, all separated by underscores.
For example, \texttt{SW3\_6.7b\_ENSV\_SV} identifies the experiment where we train the \texttt{GPT-SW3} model with 6.7 billion parameters on the bilingual English"=Swedish data and test on Swedish.

\subsection{Evaluation Methods}
\label{sec:EvaluationMethods}

\paragraph{Human Assessment}

To establish the gold standard for our evaluation, we hire three bilingual (English and Swedish) evaluators to assess model outputs based on three criteria: naturalness (how natural and fluent the generated response is), relatedness (whether the response is related to the prompt and follows the required format), and correctness (whether the response is correct, which is our main criterion). 
While there are some tasks for which these criteria may not be applicable (especially correctness), they are well"-suited for our chosen set of tasks.
We ask each annotator to rate each criterion on a Likert scale ranging from~1 (significantly deficient) to~3 (completely proficient).
For a detailed description of the annotation process and instructions, see Appendix~\ref{app:human}.
The Kendall's $\tau$ is 0.74 (averaged over pairs of annotators) and the Fleiss' $\kappa$ \cite{fleiss1973equivalence} is 0.63, indicating a substantial agreement between human annotators.

\paragraph{Majority Vote}

To compare human ratings to other metrics, we use a variant of the majority vote.
More specifically, we compress the three ratings into that score which is assigned by at least two raters, and fall back to a neutral score of~2 in cases where all raters have given different scores.
We prefer our method over the obvious alternative of taking an average because it reduces the impact of outlier ratings.
For reference, among the human ratings of correctness (our main criterion), there is a majority vote for $93.5\%$ of samples, providing a robust foundation for our comparisons.
We treat the human majority vote as the gold standard and compare other metrics against it.

\paragraph{Performance Metrics}

Our selection of performance metrics is motivated by the desire to cover commonly used methods on a spectrum from surface"-based semantics"-based methods.
On one end of this spectrum, we use \ROUGEL\ \cite{lin-2004-rouge}, which measures the textual overlap between a generated response and a reference output as the length of the longest common subsequence.
On the other end of the spectrum, we use \GPTIV\ as a judge \cite{zheng2023judging}.\footnote{Running an evaluation (across all models, tasks, and languages) costs around USD 100 and typically takes 15–20 minutes. Though not significantly cheaper than human evaluations, it certainly surpasses it in terms of time efficiency.}
Similar to previous work \cite{zhou2023lima, kew2023turning}, we prompt \GPTIV\ with the same instructions that we give to human evaluators and ask it to rate based on the same criteria on the same Likert scale.
We also prompt \GPTIV\ to provide its reasoning before rating, similarly to \citet{kew2023turning}, whose framework we use for LLM-as-a-judge evaluations.
The prompt template used for evaluations is found in Appendix~\ref{app:prompts}.
In the standard evaluation setting, the gold labels are included in the prompt as a reference for the model; we mark this setting with the suffix \texttt{-gold}.
To ablate the effects of the access to gold labels, we perform additional experiments with these labels excluded; we mark these with the suffix \texttt{-no-gold}.
Finally, as a point in"-between a purely surface"-based and a powerful semantics"-based performance metric, we use \BERTSCORE\ \cite{zhang2020bertscore}, which quantifies semantic overlap in terms of the cosine similarity between contextual embeddings obtained from pretrained language models.

\subsection{Meta-Evaluation Method}


\paragraph{Pairwise Accuracy with Tie Calibration}

In this study, we perform a meta-evaluation of both metrics that produce continuous and ordinal ratings. While Spearman's $\rho$ and Kendall's $\tau$ are commonly used for such purposes, these metrics fail to handle tasks with constant score vectors or with different rating scales, and they do not reward correct predictions of ties. Ties are especially frequent in Likert"-scale human ratings, which the automatic metrics are benchmarked against.

In response, we have chosen to use Pairwise Accuracy with Tie Calibration (\PA) for meta"-evaluating metrics. Proposed by \citet{deutsch-etal-2023-ties}, \PA\ addresses the shortcomings of traditional metrics by including mechanisms to explicitly account for the prevalence of ties, thus providing a fairer assessment of metrics.

\PA\ measures the proportion of correctly ranked pairs, including accurately predicted ties. With values ranging from 0 to 1, the metric is more easily interpreted than traditional correlation metrics such as Spearman's $\rho$ and Kendall's $\tau$. \PA\ includes a tie calibration process by defining a threshold value, $\epsilon$, which specifies what is considered a significant difference between scores. A pair of scores with a difference smaller than $\epsilon$ is considered a tie. This is crucial as some metrics are more likely to produce tied values, as can be seen in Table \ref{tab:tie_prop}. Tie calibration ensures that comparisons between different metrics are fair, regardless of their inclination to predict ties or having different rating scales.

\begin{table}[t]
\centering
\footnotesize
\begin{tabular}{lcc}
\toprule
\textbf{Metric} & \textbf{Tie Proportion} & \boldsymbol{$\epsilon$} \\
\midrule
Human Ratings & 0.557 $\pm$ 0.162 & 0.000 \\
\GPTIV\texttt{-gold} & 0.524 $\pm$ 0.154 & 0.000 \\
\ROUGEL\ & 0.355 $\pm$ 0.252 & 0.061 \\
\BERTSCORE\ & 0.104 $\pm$ 0.141 & 0.133 \\
\bottomrule
\end{tabular}
\caption{The average tie proportion per metric for English"-language tasks.}
\label{tab:tie_prop}
\end{table}

We study the distribution of ties in our data and observe significant variation for different metrics, as shown in Table \ref{tab:tie_prop}. The average tie proportion for human ratings is 0.557, serving as our benchmark. In contrast, metrics such as \GPTIV, \ROUGEL, \BERTSCORE\ exhibit varying degrees of tie proportions. \GPTIV\ has a similar degree of ties compared to human ratings, while \BERTSCORE\ has considerably lower tie proportions. 
The significant amount of ties and a constant score vector validate the use of \PA\ to enable a reliable meta-evaluation of our metrics.
As an example of a constant vector in our case, in Task 034 (cf.\ \S\ref{sec:results}), there is a constant score of 1 from human raters, illustrating a scenario that could commonly occur in instruction tuning.

Unlike \citet{deutsch-etal-2023-ties}, who calculate $\epsilon$ for each task, we calculate the optimal $\epsilon$ for each metric using data from all tasks. We find that this produces a \PA\ that better reflects the true correlation between human and metric scores, especially for tasks with only ties or mostly ties. Otherwise, for tasks with only ties, the $\epsilon$ could be as large as the value range for the metric and treat every pair of scores as ties. As shown in Table \ref{tab:tie_prop}, our metric-level $\epsilon$ correlates well with the number of ties for the metric.
With our pre-calculated $\epsilon$ values, we compute \PA\ over all models per task, aligning with the No-Grouping setting in \citet{deutsch-etal-2023-ties}. 

\section{Results and Analysis}
\label{sec:results}

In this section, we present a comparative analysis of the evaluation methods in terms of their alignment with human assessments.

\subsection{Human Evaluation}

\begin{table}[t]
\centering
\footnotesize
\begin{tabular}{lccc}
\toprule
\textbf{Experiment} & \bf Natural & \bf Related & \bf Correct \\
\midrule
\multicolumn{4}{l}{\bf{}\texttt{LLaMA2\_13b}} \\
\quad\texttt{EN\_EN} & $99$ \tiny$(95)$ & $73$ \tiny$(80)$ & $47$ \tiny$(51)$ \\
\quad\texttt{SV\_SV} & $99$ \tiny$(94)$ & $76$ \tiny$(76)$ & $36$ \tiny$(37)$ \\
\quad\texttt{ENSV\_EN} & $99$ \tiny$(92)$ & $80$ \tiny$(79)$ & $47$ \tiny$(47)$ \\
\quad\texttt{ENSV\_SV} & $99$ \tiny$(92)$ & $80$ \tiny$(76)$ & $39$ \tiny$(40)$ \\
\midrule
\multicolumn{4}{l}{\bf{}\texttt{LLaMA2\_7b}} \\
\quad\texttt{EN\_EN} & $99$ \tiny$(91)$ & $72$ \tiny$(76)$ & $40$ \tiny$(44)$ \\
\quad\texttt{SV\_SV} & $99$ \tiny$(88)$ & $65$ \tiny$(68)$ & $35$ \tiny$(34)$ \\
\quad\texttt{ENSV\_EN} & $99$ \tiny$(92)$ & $74$ \tiny$(75)$ & $41$ \tiny$(43)$ \\
\quad\texttt{ENSV\_SV} & $98$ \tiny$(89)$ & $66$ \tiny$(68)$ & $32$ \tiny$(33)$ \\
\midrule
\multicolumn{4}{l}{\bf{}\texttt{SW3\_6.7b}} \\
\quad\texttt{EN\_EN} & $99$ \tiny$(91)$ & $79$ \tiny$(70)$ & $35$ \tiny$(36)$ \\
\quad\texttt{SV\_SV} & $99$ \tiny$(92)$ & $66$ \tiny$(67)$ & $37$ \tiny$(37)$ \\
\quad\texttt{ENSV\_EN} & $99$ \tiny$(92)$ & $81$ \tiny$(68)$ & $35$ \tiny$(35)$ \\
\quad\texttt{ENSV\_SV} & $99$ \tiny$(85)$ & $68$ \tiny$(63)$ & $35$ \tiny$(36)$ \\
\midrule
\textbf{Avg.\ diff.} & $7.8$ & $4.2$ & $1.3$ \\
\bottomrule
\end{tabular}
\caption{Human evaluation results per model scaled to $0$ to $100$. For comparison, \GPTIV\ ratings are included in parentheses for each model and criterion.}
\label{tab:acc}
\end{table}

We present the human evaluation results for each model in Table~\ref{tab:acc}.
All models demonstrate the capability to generate natural"-sounding text (99\% on average) and also perform fairly well in generating relevant responses that adhere to the required format (73\% on average).
The correctness scores demonstrate that the models are capable of generating largely accurate answers.
There is also a notable diversity among the models regarding correctness.
This diversity is crucial because we seek a range of models with varied problem"-solving abilities, rather than just strong models that produce highly accurate results.

The ratings of \GPTIV\ closely align with human ratings for correctness, showing slightly more distance in relatedness, and even more in naturalness.
(The average differences between human and \GPTIV\ ratings are summarized in the final row of Table~\ref{tab:acc}.)
Based on these results, in the rest of the paper, we focus solely on correctness.
We prioritize correctness since it is the most important criterion for determining the usefulness of LLMs. Moreover, comparing our metrics using the other criteria could be problematic. For instance, while \ROUGEL\ scores serve as a reasonable proxy for correctness, they are less suited for evaluating naturalness or relatedness.



\subsection{Meta-Evaluation of Metrics}
The metric with the highest alignment with human ratings is \GPTIV\texttt{-gold} which achieves an average \PA\ of 0.81 for English short- and long-answer tasks, followed by \ROUGEL\ with 0.75, \BERTSCORE\ with 0.66, and \GPTIV\texttt{-no-gold} with 0.62. For a comparison of all the results across different task types, languages, and metrics, see Table \ref{tab:metrics}. Nonetheless, a more fine-grained analysis shows a different pattern, which we will discuss in this section.



\begin{figure*}[ht]
    \centering
    \includegraphics[width=\textwidth]{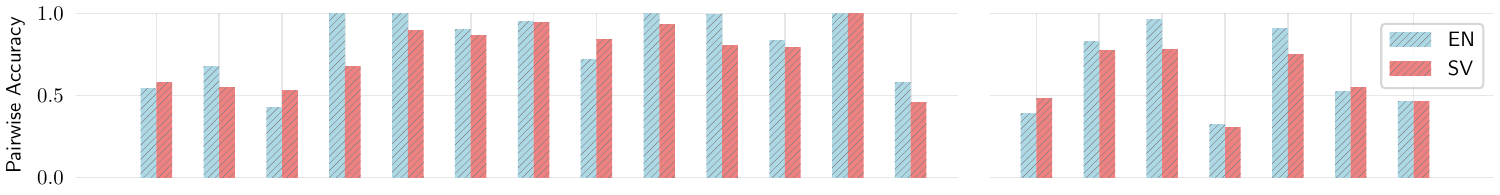}   
    \includegraphics[width=\textwidth]{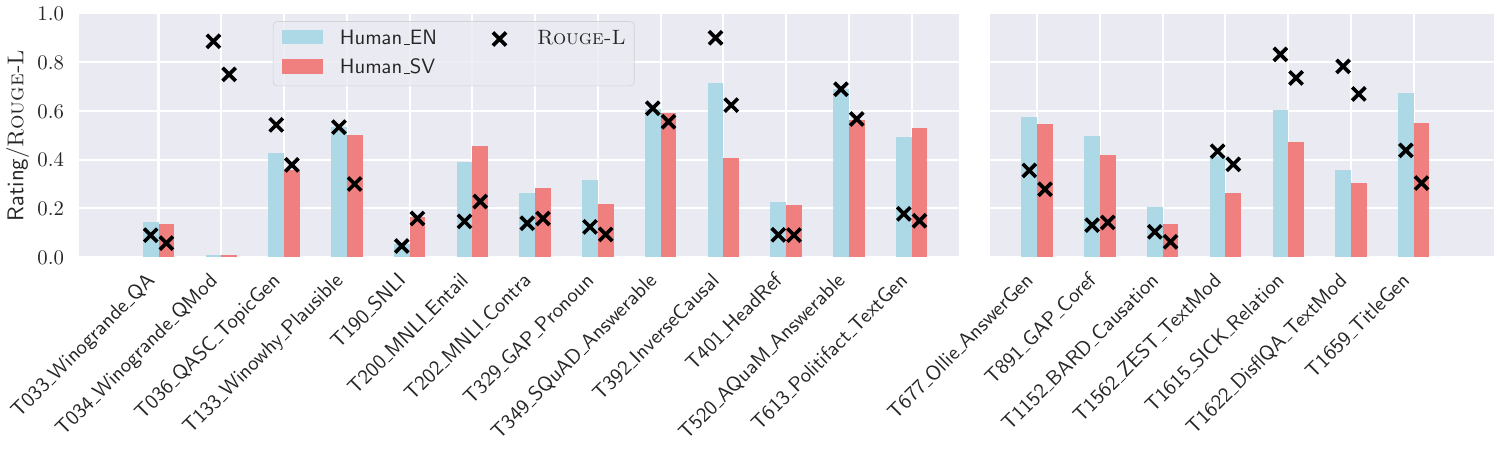}
    \caption{Human ratings and \ROUGEL\ scores per task and test language at the bottom and their \PA\ at the top. Human scores are normalized to a range between 0 and~1. Short-answer tasks are on the left and long-answer ones are on the right.}
    \label{fig:rouge}
\end{figure*}

\begin{finding}{1}
All metrics struggle to assess model performance on long-answer tasks.
\end{finding}

\begin{figure*}[ht]
    \centering
    \includegraphics[width=\textwidth]{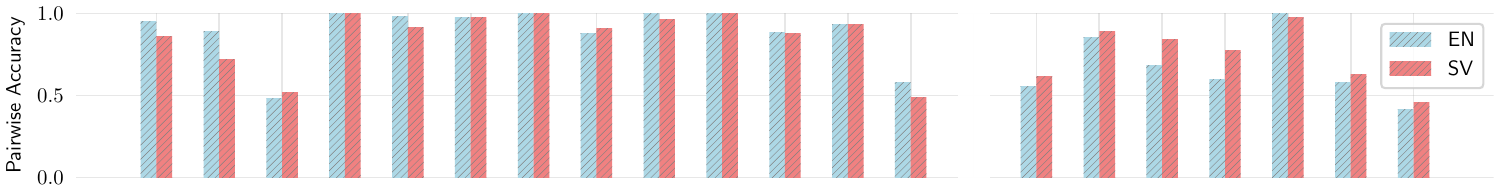}
    \includegraphics[width=\textwidth]{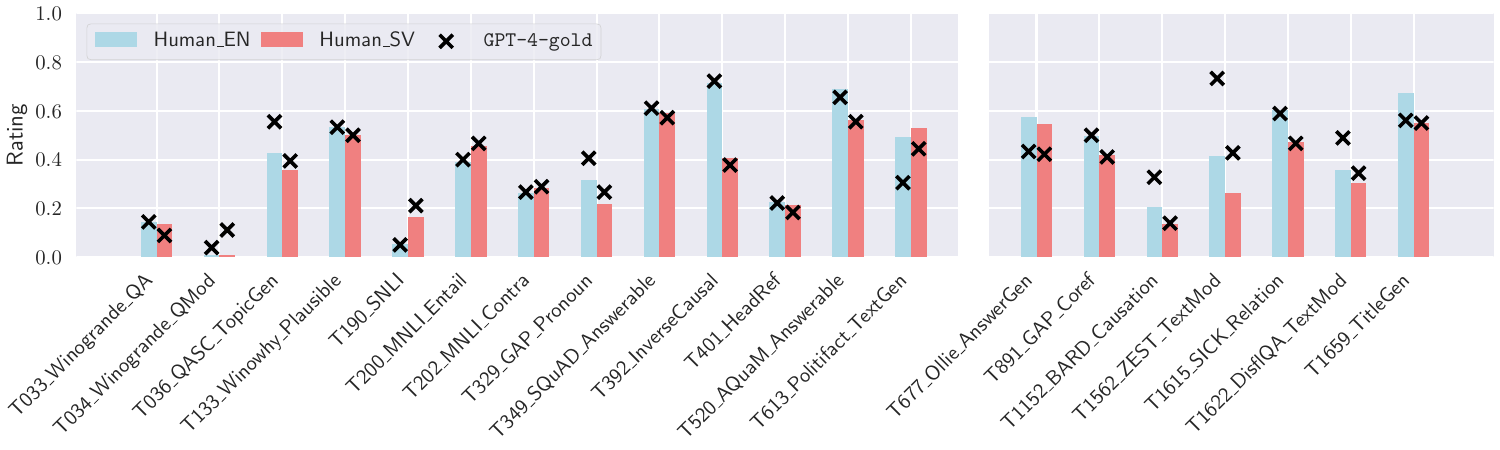}
    \caption{Human and \GPTIV\texttt{-gold}'s ratings per task and test language on the bottom and their \PA\ on top. Ratings are normalized to a range of 0 to 1. Short-answer tasks are on left and long-answer ones on right.}
    \label{fig:gpt-4-gold}
\end{figure*}

\begin{figure*}[ht]
    \centering
    \includegraphics[width=\textwidth]{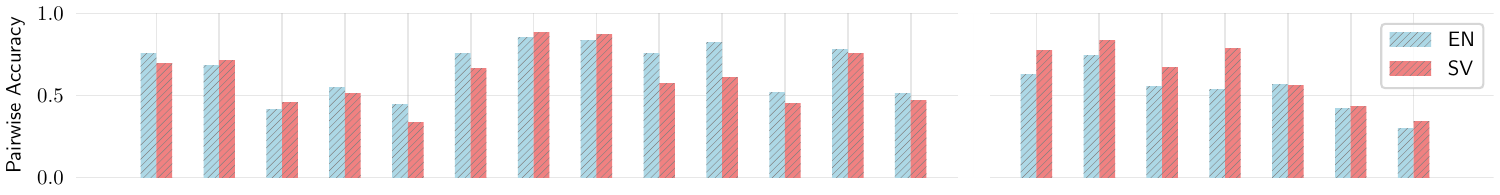}
    \includegraphics[width=\textwidth]{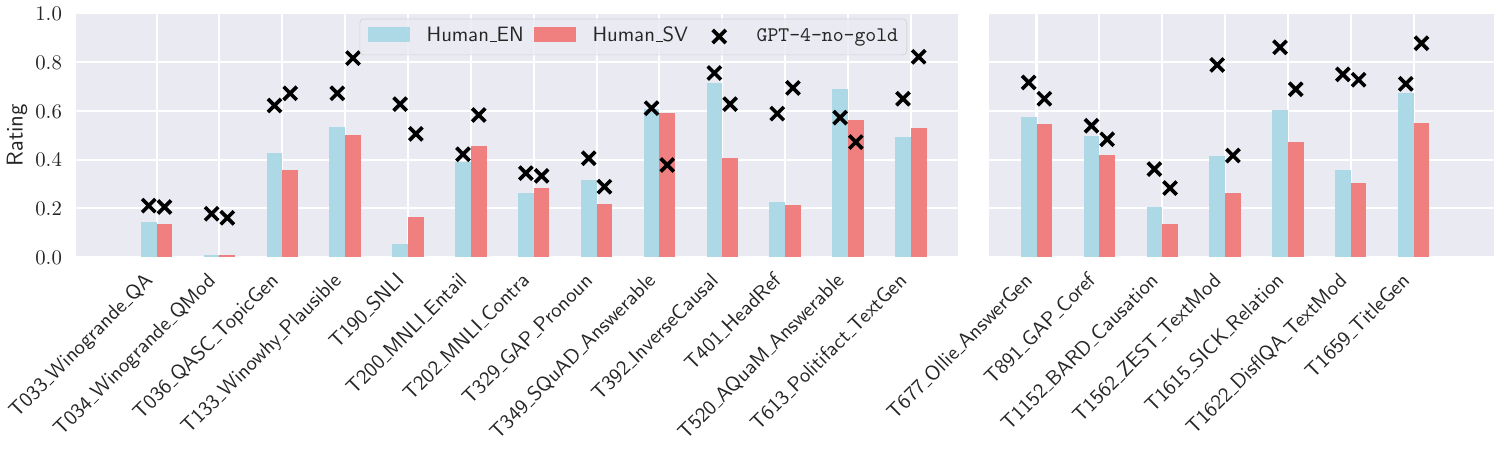}
    \caption{Human and \texttt{GPT-4-no-gold}'s ratings per task and test language on the bottom and their \PA\ on top. Ratings are normalized to a range of 0 to 1. Short-answer tasks are on left and long-answer ones on right.}
    \label{fig:gpt-4-nogold}
\end{figure*}

As presented in Figures \ref{fig:rouge}, \ref{fig:gpt-4-gold}, and \ref{fig:gpt-4-nogold}, the alignment with humans drops for long-answer tasks compared to short-answer tasks. On average, for English long-answer tasks, \GPTIV\texttt{-gold} at 0.58 is the highest, followed by 0.54 for \BERTSCORE, 0.50 for \GPTIV\texttt{-no-gold}, and 0.48 for \ROUGEL.

For Tasks 613, 677 and 1659 where the models must generate free-form text, \ROUGEL\ scores are lower than human ratings accompanied with low to medium \PA s.
This is due to the large set of possible solutions and the small set of gold label answers, often consisting of a single sentence, that the output should match to receive a high \ROUGEL\ score. 
For some of these tasks, e.g., title generation, it is impossible to cover the set of conceivable solutions which makes \ROUGEL\ unreliable for these types of tasks.
The same trend is present for \GPTIV, which could be explained by this model's relying on the gold labels during evaluation, which may make it overly critical of responses that show less conformity with the provided reference answers.

In contrast, for Tasks 1562 and 1622, where models are tasked with modifying sentences, \GPTIV\ assigns higher ratings compared to humans.
\GPTIV\ struggles to reliably assess whether our models have met the instructional criteria in such tasks.
For instance, in Task 1562, the objective is to generate paraphrases of questions while making as many alterations as possible.
When models only introduce minor changes, such as changing \textit{bring} to \textit{take} in the sentence ``Can I bring my mountain bike with me to this national park?'', \GPTIV\ often rates this 3 on correctness.

\begin{figure}[th]
    \centering
    \includegraphics[width=\columnwidth]{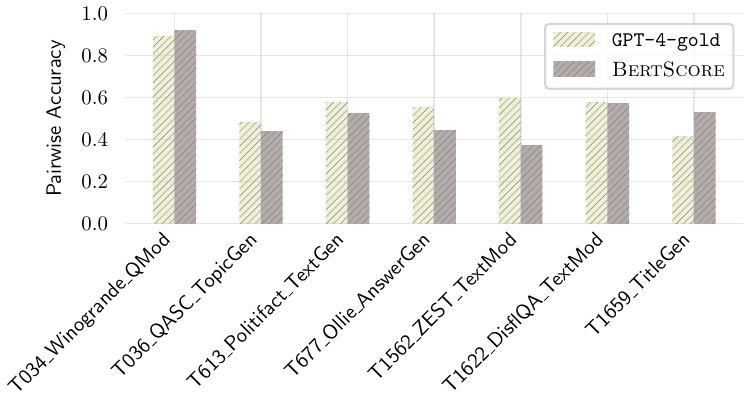}
    \caption{Pairwise accuracy between \texttt{GPT-4-gold} and \BERTSCORE\ for long-answer English tasks.}
    \label{fig:bvsg}
\end{figure}

\begin{finding}{2}
\GPTIV\ needs reference answers in the prompt.
\end{finding}
When we remove the gold answers from the prompt, \GPTIV's alignment with human ratings decreases significantly, from 0.81 \PA\ to 0.62 \PA\ on English. Full results are presented in Table \ref{tab:metrics}.
For short"-answer tasks in English, the reduction is 0.25, and for long-answer tasks, it is 0.09, which underscores \GPTIV's struggle when it lacks a reference for exact matching.
We attribute this reduction to the increased complexity of both solving and rating the task, which is more pronounced in Swedish, where there is a higher incidence of models over-generating.
While such over-generation may lead to lower human ratings, it could be favored by \GPTIV\ due to its bias towards longer and more verbose outputs \cite{zheng2023judging}.

A notable observation is that without gold label references, \GPTIV\texttt{-no-gold} is generally more prone to higher ratings, as seen in Figure \ref{fig:gpt-4-nogold}'s lower plot.
While \GPTIV\texttt{-gold} closely aligns with human judgments for incorrect outputs (88\% for 1's and 81\% for 3's), \GPTIV\texttt{-no-gold} shows an alignment of 65\% for 1's and 84\% for 3's, indicating an opposite trend. This demonstrates the excessively positive stance of \GPTIV\texttt{-no-gold}, also noted by \citet{hada2024large} in a different scenario.
The positive bias is particularly evident in long-answer tasks and challenging short-answer tasks, such as Tasks 190 and 401.
This tendency underscores the importance of gold labels as references to help align the \GPTIV's judgments with humans in most tasks.

However, we note that for some long generation tasks (Tasks 613, 677, and 1659) the scores of human raters are lower than for other tasks even when gold labels are present, as seen in Figure \ref{fig:gpt-4-gold}.
Gold references can therefore be restricting for these types of long generation tasks, where models can have correct answers that diverge from the gold label.
This is problematic as it is to these types of tasks that LLM-as-a-judge is often applied, and where it could bring the most value compared to other metrics, particularly due to the multitude of potential correct answers.


\begin{finding}{3}
For short-answer tasks, \ROUGEL\ is as effective as \GPTIV\texttt{-gold}.
\end{finding}

With a \PA\ at 0.90 for English short-answer tasks, \ROUGEL\ is nearly as well"-aligned with human ratings as \GPTIV\texttt{-gold}, which has a \PA\ at 0.93.
Our strong results for \ROUGEL\ are in line with findings by \citet{wang2022super}, who report a high correlation with humans for classification tasks. 
With that said, there are instances where \ROUGEL\ has low alignment with human majority rating for short-answer tasks, for example when there is high word overlap between possible answer choices.
For example, in Task 1615 the labels are ``B entails A'', ``B contradicts A'', or ``B neutral A''.
A wrong answer for this task yields a \ROUGEL\ score of 0.66, as long as the answer is in the possible answers space.
The same issue is observed for Task 392 where the label space is ``Plausible'' and ``Not Plausible''.

These types of issues also make it problematic to report average \ROUGEL\ scores across tasks since a baseline model that always makes wrong predictions could inflate its score beyond its actual performance level.
However, as previously discussed, \ROUGEL\ correlation with humans and \GPTIV\texttt{-gold} does not carry over to long-answer tasks, which makes it only suitable as a replacement for \GPTIV\ when evaluating short-answer tasks.
It is important to note that while \ROUGEL\ demonstrates strong agreement with humans, \GPTIV\texttt{-gold} scores are more interpretable as they better align with human judgments, as illustrated at the bottom of Figures \ref{fig:rouge} and \ref{fig:gpt-4-gold}.



\begin{finding}{4}
\BERTSCORE\ demonstrates strong performance in long-answer tasks.
\end{finding}

With a \PA\ of 0.54 for long-answer tasks, \BERTSCORE\ shows a alignment with humans comparable with \GPTIV\texttt{-gold}, which scores 0.58. A comparison of \BERTSCORE\ and \GPTIV\texttt{-gold}'s \PA s are shown in Figure \ref{fig:bvsg}. For complete results of \BERTSCORE\ see Appendix \ref{app:bertscore}.
\BERTSCORE\ achieves comparable results to \GPTIV\texttt{-gold} on all long-answer tasks, only underperforming on some of them, particularly Task 1622, where it captures the first criterion which requires a high semantic similarity between the two, but fails to take into account whether enough words have been changed, such as in cases involving synonyms, which is explicitly mentioned in the instructions.


\begin{finding}{5}
Swedish presents a challenge for certain metrics.
\end{finding}

For \ROUGEL, a reduction of 0.074 in \PA\ is observed for Swedish compared to English. In contrast, \GPTIV\texttt{-gold} experiences no significant reduction when switching from English to Swedish. However, \GPTIV\texttt{-no-gold} is less consistent, showing a reduction of 0.044.
\GPTIV's decrease in performance when it does not have access to gold references for Swedish outputs suggests that the model has more difficulties solving the task when it is in another language.
We believe the effect could be even more pronounced for languages with less training data than Swedish or less typologically similar to English.
Further studies would be necessary to investigate this hypothesis.

While \ROUGEL\ does not take language into account, it may be less reliable a measurement for languages other than English due to models' failing to adhere to the required format. 
For instance, we observe that Swedish models have difficulties following instructions, even for short-answer tasks.
They sometimes generate synonyms to the true labels, e.g., \textit{sannolikt} ``probable'' instead of \textit{troligt} ``likely'', an effect that could stem from seeing less data in Swedish and therefore having less reliable instruction-following capabilities.
This is particularly concerning given the prevalence of work utilizing automatic evaluation measures across different languages.

%

\section{Conclusions and Future Work}

This study provides insights into the methods we use to evaluate language model generations, focusing on when automatic metrics align with human annotators and what the best metric is under different scenarios. We are the first to do a broader meta-evaluation study where we compare \GPTIV-as-a-judge and traditional metrics with a methodology that allows for reliable comparisons between metrics.
We recommend using Pairwise Accuracy (PA) with Tie Calibration for meta-evaluation. This method effectively handles ties, which are prevalent when using human and \GPTIV\ ratings, making it a reliable tool for assessing metric performance against human ratings.

Our main finding is that \GPTIV\ shows strong alignment with human judgments for short-answer tasks, but only when gold references are provided. The reliability drops significantly without gold references, as the model is overly positive compared to human evaluations. The issue is particularly evident in free-form tasks, which are tasks where LLM-as-a-judge could be the most valuable and where gold labels are typically not available. When gold references are available, we observe that \GPTIV\ is too strict compared to humans, relying to much on the gold label. For these type of tasks, even though LLM-as-a-judge is often applied to them, human evaluations still remain the gold standard.
 
\ROUGEL\ performs comparably to \GPTIV\texttt{-gold} for short-answer tasks, offering a cost-effective alternative in scenarios where the use of \GPTIV\ is limited by cost or time constraints.
For long-answer tasks, while \BERTSCORE\ demonstrates strong performance, it does not fully replace the need for \GPTIV\texttt{-gold}. These metrics provide valuable insights but vary in effectiveness depending on the specific task.

Evaluating non-English outputs, such as Swedish, presents additional challenges. There is a significant drop in alignment for \GPTIV\texttt{-no-gold}, which highlights that \GPTIV\ as a judge is less reliable for languages other than English. While this is true for Swedish, we expect these findings to be more sever for lesser-resourced languages and those less similar to English. Future work should focus on expanding this study to more languages.

As we have observed, there is a large variation in alignment with human ratings for all metrics across task types. Previous research identifies strong correlations with human annotators, but that is often the average over tasks. Our findings underscores the necessity of task-specific evaluation metrics rather than relying on general averages which can obscure important nuances in metric alignment with human annotators.
Furthermore, while \GPTIV\ is a valuable tool for evaluation short-answer tasks when gold references are available, alternatives like \ROUGEL\ and \BERTSCORE\ can be effective for most tasks types, offering cost-efficient and reliable evaluations.


\section*{Limitations}
We choose to report our results using pairwise accuracy which we believe provides more robust and reliable alignment statistics compared to common correlation metrics. With that said, \PA\ has its shortcomings, such as when it faces constant or close-to-constant scores. For example, when the reference vector is $\vec{1}$, a metric vector of $\vec{1}$ or $\vec{3}$ both result in very high \PA s, due to the lack of prior knowledge about the metric range. However, this issue is not unique to \PA; common correlation metrics also face the same challenge in the case of close-to-constant vectors.

Our study primarily focuses on the evaluation of tasks across English and Swedish. Consequently, the findings may not be applicable to languages that have syntax and structure significantly different from English. We deliberately made this choice to enable a broader examination of tasks and tap into expert knowledge for deeper analyses. Essentially, we prioritized expanding the range of tasks and delving deeper into analysis rather than focusing on additional languages.
Furthermore, our evaluation exclusively uses \GPTIV\ as the language model for assessment. The rapidly evolving landscape of language models suggests the existence of other models that may yield different results or exhibit different patterns.

\section*{Ethical Considerations}
Our annotators, residents of Sweden, were selected for their proficiency in both English and Swedish, ensuring precise interpretation and annotation of content. We ensured their fair compensation in line with prevailing norms for similar tasks in Sweden. Furthermore, they completed their assignments within a reasonable timeframe, enabling them to work without undue pressure. Prior to acceptance, annotators were briefed on the purpose of their annotations ensuring that they understood the objectives and context behind the tasks assigned to them.

\section*{Acknowledgments}
We extend our gratitude to our annotators for their diligent contributions to this paper and to Daniel Deutsch and Richard Johansson for their valuable feedback.
This work was partially supported by the Wallenberg AI, Autonomous Systems and Software Program (WASP), funded by the Knut and Alice Wallenberg Foundation and by the European Union’s Horizon 2020 research and innovation program under grant agreement No 101135671 (TrustLLM).
We also acknowledge support from the National Graduate School of Computer Science in Sweden (CUGS).
The computations were enabled by the Berzelius resource provided by the Knut and Alice Wallenberg Foundation at the Swedish National Supercomputer Center.

\bibliography{anthology,custom}
\bibliographystyle{acl_natbib}

\newpage

\appendix
\renewcommand{\thefigure}{A\arabic{figure}} 
\renewcommand{\thetable}{A\arabic{table}}   
\setcounter{table}{0}  
\setcounter{figure}{0} 

\section{Task Descriptions}
\label{app:tasks}
\paragraph{Task033: Winogrande Answer Generation}
A fill-in-the-blank task with some restriction, such as that the answer should be chosen from the two objects in the question. "I planted more tomato seeds than I planted cucumber seeds since I hated eating the \_ ." Gold answer: "cucumber".

\paragraph{Task034: Winogrande Question Modification Object}
Similar to task033, but this time the task is to change the question so that the answer, which is given in the input, changes to the other object present in the input.

\paragraph{Task036: QASC Topic Word to Generate Related Fact}
Write a topic word for the given fact with at least one word overlap with the fact. Example: "Fact: a seismograph is used for measuring the size of an earthquake." One possible gold answer: "seismograph earthquake."

\paragraph{Task133: Winowhy Reason Plausibility Detection}
Indicate the plausibility of reasoning for the pronoun coreference relations. Example: "Sentence: Although they ran at about the same speed, Sue beat Sally because she had such a bad start.\textbackslash{}n Reason: The 'she' refers to sally because Sue won, sally lost. \textbackslash{}n Question: Is the above reasoning correct or wrong? " Gold answer: "Correct". 

\paragraph{Task190: SNLI Classification}
Given two sentences, classify their agreement: entailment, contradiction, or neutral.

\paragraph{Task200: MNLI Entailment Classification}
From three options, choose the one that can be inferred from the given sentence. 

\paragraph{Task202: MNLI Contradiction Classification}
From three options, choose the one that disagrees with the given sentence. 

\paragraph{Task329: GAP Classification}
Given a text, a pronoun, and two candidate names, determine which of the names the pronoun refers to. The answer should be either A, B, or neither.

\paragraph{Task349: SQuAD2.0: Answerable Unanswerable Question Classification}
Determine whether or not the given question is answerable by the provided passage.

\paragraph{Task392: Inverse Causal Relationship}
Given two sentences separated by the word "because", determine whether the second sentence can be the result of the first one (is there a cause and effect relationship?)

\paragraph{Task401: Numeric Fused Head Reference}
Using your knowledge about language and commonsense, determine what element the marked number refers to. Example: "Jim Bronson:  What 's your name ?\textbackslash{}nTemple Brooks: I do n't have \_ one \_ !\textbackslash{}nJim Bronson: Well everyone I have ever know had a name , that 's really weird . My name is Jim incase your interested .\textbackslash{}nTemple Brooks: Well I 'm not !" Gold answer: "name".

\paragraph{Task520: AQuaMuSe Answer Given in Passage}
Is the answer to the given question contained in the provided passage?

\paragraph{Task613: PolitiFact Text Generation}
Generate the subject of a speech by a politician.

\paragraph{Task677: Ollie Sentence Answer Generation}
Given two noun phrases (arguments) and the relationship between them, write a sentence that expresses theses arguments with the given relationship. 

\paragraph{Task891: GAP Coreference Resolution}
Given a passage, find the corresponding person for the provided pronoun. 

\paragraph{Task1152: BARD Analogical Reasoning Causation}
Replace question mark with a verb which is the appropriate consequence of the given action. For example: "ignite : burn. hit : ?". Gold answer: "shatter".

\paragraph{Task1562: ZEST Text Modification}
Paraphrase the given questions to have different wording. Change it as much as possible using synonyms, etc. Example: "Does this dog breed always have spots?". 

\paragraph{Task1615: SICK Classify b Relation a}
Classify the relation between two sentences: B\_entails\_A, B\_contradicts\_A, or B\_neutral\_A.

\paragraph{Task1622: Disfl-QA: Text Modification}
Convert a disfluent question to a proper question. Example: "Who were uh instead tell me how many quadrangles does the Main Quadrangles have?"

\paragraph{Task1659: Title Generation}
Generate a title under forty words which mentions the purpose of the text.

\section{Prompt Templates}
\label{app:prompts}

\begin{figure*}[ht]
    \centering 
    \includegraphics[width=\textwidth]{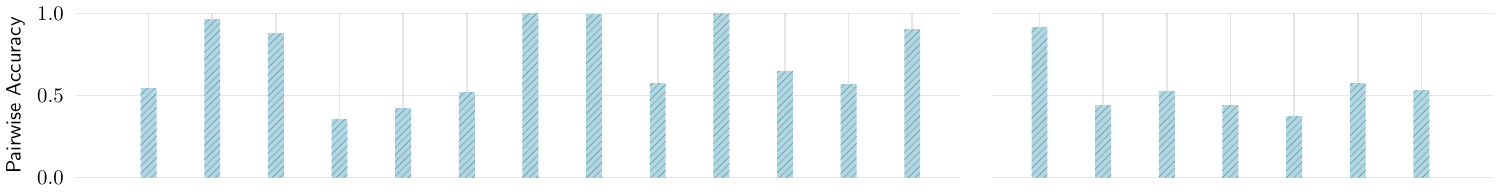}   
    \includegraphics[width=\textwidth]{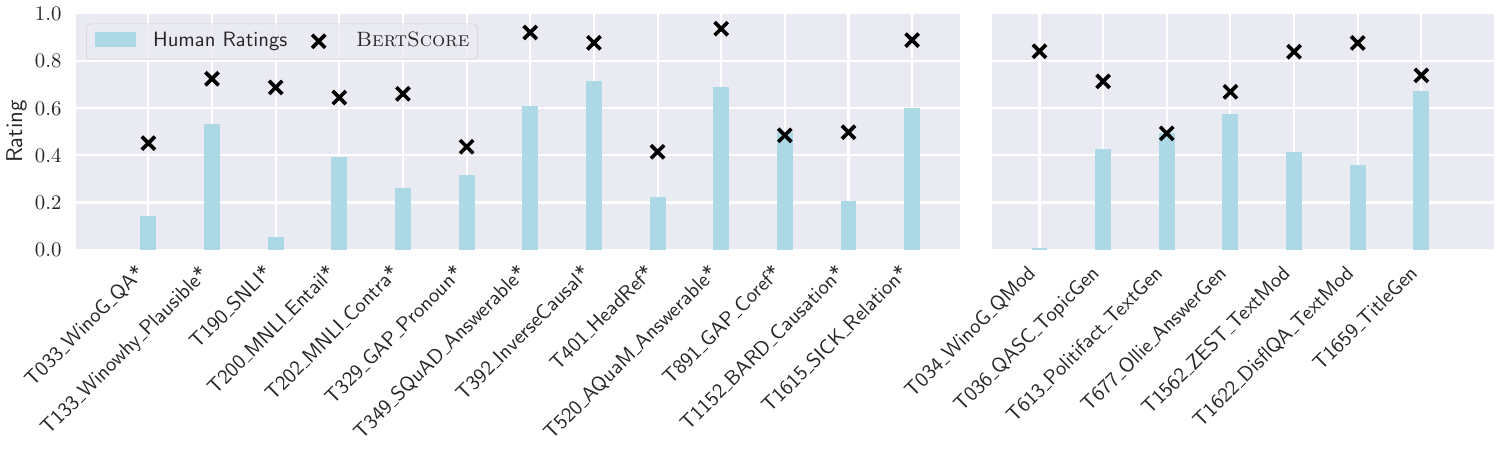}
    \caption{Human and \BERTSCORE\ ratings per task and test language on the bottom and their \PA\ on top. The human ratings are normalized to a range of 0 to 1. Short-answer tasks are on left and long-answer ones on right.}
    \label{fig:bertscore}
\end{figure*}

\subsection{Translation}
We use the following prompt for translating our datasets from English to Swedish using \texttt{GPT-3.5-turbo}:

\begin{lstlisting} 
Translate the following text from
English to Swedish:
{English text}
\end{lstlisting}

\subsection{LLM-as-a-judge}
The prompt used for \GPTIV\ could be found in Table \ref{tab:gpt4prompt}. The prompt for \GPTIV\texttt{-no-gold} is the same, but without the following part:

\begin{lstlisting}
[Gold Answer] (If there are several gold answers then they are all correct alternatives):  {gold_answer}
***
\end{lstlisting}

\section{Supplementary Results}

\subsection{How Good Are Our Language Models at Swedish?}
\label{app:ppl}
To assess the effectiveness of the models described in Section \ref{sec:method}, we measure their perplexity. To ensure the generated texts meet high standards and to avoid assessing the models on data used during their pretraining, we use a custom dataset consisting of current news articles from SVT\footnote{\href{https://www.svt.se/}{svt.se}}, the Swedish national public television broadcaster. The dataset comprises 268 articles spanning various topics, published between June 1st, 2023, and October 16th, 2023.
To address the variability caused by different tokenizers across various models, we use character length normalization when calculating perplexity \citep{liang2022holistic, yong-etal-2023-bloom}.

\begin{table}[t]
\centering
\begin{tabular}{lc}
\toprule
\textbf{Model} & \textbf{Perplexity} \\
\midrule
\texttt{LLaMA2\_13b} & 1.96 
\\
\texttt{LLaMA2\_13b\_EN} & 2.03  
\\
\texttt{LLaMA2\_13b\_SV} & 2.27  
\\
\texttt{LLaMA2\_13b\_ENSV} & 2.24  
\\
\midrule
\texttt{LLaMA2\_7b} & 2.09  
\\
\texttt{LLaMA2\_7b\_EN} & 2.22 
\\
\texttt{LLaMA2\_7b\_SV} & 2.51  
\\
\texttt{LLaMA2\_7b\_ENSV} & 2.49  
\\
\midrule
\texttt{SW3\_6.7b} & 1.62  
\\
\texttt{SW3\_6.7b\_EN} & 1.65  
\\
\texttt{SW3\_6.7b\_SV} & 1.72  
\\
\texttt{SW3\_6.7b\_ENSV} & 1.69  
\\
\bottomrule
\end{tabular}
\caption{The perplexity of our models on the SVT dataset. The abbreviations are the training language(s).
}
\end{table}

The perplexity for Swedish consistently remains lower in models prior to instruction tuning. However, after tuning, the poorest outcomes are noted in the \texttt{SV} models trained solely on Swedish data. Interestingly, the \texttt{ENSV} models exhibit improved performance, with the \texttt{EN} models showing even better results. This variation could be ascribed to the Swedish-specific model weights being less affected due to their lower exposure to Swedish data, but requires further investigations. Notably, the increments in perplexity are less pronounced for the \texttt{SW3} models.

\subsection{Pairwise Accuracy per Model}
\label{app:rouge}
\label{app:gpt4}
\label{app:gpt4-no-gold}
Pairwise accuracy per model is shown in Table \ref{tab:permodel}.

\begin{table*}[!h]
\centering
\begin{tabular}{lccccc}
\toprule
\textbf{Model} & \textbf{\GPTIV\texttt{-gold}} & \textbf{\GPTIV\texttt{-no-gold}} & \textbf{\ROUGEL} & \textbf{\BERTSCORE} \\
\midrule
\texttt{LLAMA2\_13b\_EN\_EN} & 0.779 & 0.590 & 0.601 & 0.512 \\
\texttt{LLAMA2\_13b\_SV\_SV} & 0.816 & 0.559 & 0.547 & - \\
\texttt{LLAMA2\_13b\_ENSV\_EN} & 0.782 & 0.594 & 0.639 & 0.545 \\
\texttt{LLAMA2\_13b\_ENSV\_SV} & 0.828 & 0.565 & 0.634 & - \\
\midrule
\texttt{LLAMA2\_7b\_EN\_EN} & 0.766 & 0.581 & 0.578 & 0.454 \\
\texttt{LLAMA2\_7b\_SV\_SV} & 0.794 & 0.578 & 0.526 & - \\
\texttt{LLAMA2\_7b\_ENSV\_EN} & 0.794 & 0.609 & 0.607 & 0.447 \\
\texttt{LLAMA2\_7b\_ENSV\_SV} & 0.804 & 0.577 & 0.550 & - \\
\midrule
\texttt{SW3\_6.7b\_EN\_EN} & 0.789 & 0.612 & 0.604 & 0.482 \\
\texttt{SW3\_6.7b\_SV\_SV} & 0.843 & 0.645 & 0.505 & - \\
\texttt{SW3\_6.7b\_ENSV\_EN} & 0.830 & 0.606 & 0.622 & 0.491 \\
\texttt{SW3\_6.7b\_ENSV\_SV} & 0.828 & 0.576 & 0.600 & - \\
\bottomrule
\end{tabular}
\caption{Pairwise accuracy per model for all metrics.}
\label{tab:permodel}
\end{table*}

\subsection{\BERTSCORE\ Results}
\label{app:bertscore}
Detailed results of \BERTSCORE\ are shown in Figure \ref{fig:bertscore}.

\begin{table*}[h!]
\centering
\begin{tabular}{lccccccc}
\toprule
\textbf{Model Name} & \textbf{Task Type} & \textbf{Language} & \boldsymbol{$\tau$} & \boldsymbol{$\rho$} & \boldsymbol{\PA} & \boldsymbol{$\epsilon$} \\
\midrule
\ROUGEL\ & all & EN & 0.667 & 0.712 & 0.752 & 0.061 \\
\ROUGEL\ & long & EN & 0.308 & 0.368 & 0.484 & 0.061 \\
\ROUGEL\ & short & EN & 0.833 & 0.871 & 0.896 & 0.061 \\
\ROUGEL\ & all & SV & 0.584 & 0.631 & 0.699 & 0.095 \\
\ROUGEL\ & long & SV & 0.325 & 0.389 & 0.477 & 0.095 \\
\ROUGEL\ & short & SV & 0.704 & 0.743 & 0.818 & 0.095 \\
\midrule
\GPTIV\texttt{-gold} & all & EN & 0.781 & 0.801 & 0.811 & 0.000 \\
\GPTIV\texttt{-gold} & long & EN & 0.481 & 0.521 & 0.584 & 0.000 \\
\GPTIV\texttt{-gold} & short & EN & 0.920 & 0.930 & 0.934 & 0.000 \\
\GPTIV\texttt{-gold} & all & SV & 0.792 & 0.811 & 0.817 & 0.000 \\
\GPTIV\texttt{-gold} & long & SV & 0.538 & 0.575 & 0.601 & 0.000 \\
\GPTIV\texttt{-gold} & short & SV & 0.909 & 0.920 & 0.934 & 0.000 \\
\midrule
\GPTIV\texttt{-no-gold} & all & EN & 0.517 & 0.535 & 0.622 & 0.000 \\
\GPTIV\texttt{-no-gold} & long & EN & 0.309 & 0.338 & 0.499 & 0.000 \\
\GPTIV\texttt{-no-gold} & short & EN & 0.613 & 0.625 & 0.688 & 0.000 \\
\GPTIV\texttt{-no-gold} & all & SV & 0.496 & 0.514 & 0.620 & 0.000 \\
\GPTIV\texttt{-no-gold} & long & SV & 0.393 & 0.425 & 0.569 & 0.000 \\
\GPTIV\texttt{-no-gold} & short & SV & 0.544 & 0.556 & 0.648 & 0.000 \\
\midrule
\BERTSCORE\ & all & EN & 0.419 & 0.482 & 0.658 & 0.133 \\
\BERTSCORE\ & long & EN & 0.367 & 0.452 & 0.542 & 0.133 \\
\BERTSCORE\ & short & EN & 0.443 & 0.495 & 0.720 & 0.133 \\
\bottomrule
\end{tabular}
\caption{Comparison of metrics across different task types for English and Swedish. $\tau$ stands for Kendall's and $\rho$ for Spearman's rank correlation coefficient.}
\label{tab:metrics}
\end{table*}

\section{Implementation Details}
\label{app:imp}
Following \citet{alpaca}, we finetune \texttt{LLAMA2\_7b} and \texttt{SW3\_6.7b}, which is roughly the same size, for 3 epochs and with a learning rate of $2e-5$, and the larger \texttt{LLAMA2\_13b} model for 5 epochs and with a learning rate of $1e-5$. The batch size is set to 128 for both cases. Unlike \citet{alpaca}, we allow for a longer maximum length of 2048 and truncate longer samples. For the sake of reducing computational costs we opt for using \texttt{bf16} and \texttt{tf32} precision formats. We distribute the training across multiple GPUs using DeepSpeed \cite{10.1145/3394486.3406703} stage 3 without offloading. 

\section{Human Evaluations}
\label{app:human}
Our evaluators are not crowd-sourced workers; instead, they are individuals with some experience and expertise in the field. They were carefully selected for their familiarity with the subject matter and were hired specifically for this evaluation task. While they were instructed to use their own judgment in the assessment, they also had access to the gold standard answers to guide their evaluations. The instructions given to human evaluators were similar to those given to \GPTIV, as presented in Figure \ref{fig:bvsg}.

\begin{figure*}[b]
\begin{lstlisting}
system\_message = """You are an expert language evaluator."""

user\_message = """You are evaluating a response that has been submitted for a particular task, using a specific set of standards. Below is the data: 

[BEGIN DATA]
***
[Task]: {prompt} 
***
[Submission]: {response}
***
[Gold Answer] (If there are several gold answers then they are all correct alternatives):  {gold_answer}
***
[Criterion]: Evaluation Criteria

Naturalness:
1: "Not at all natural - The generated text is grammatically incorrect or sounds unnatural, including awkward phrasing or inappropriate vocabulary."
2: "Somewhat natural - The generated text has minor grammatical errors or slightly awkward phrasing but is mostly understandable and natural."
3: "Completely natural - The generated text is grammatically correct, well-phrased, and uses appropriate vocabulary, sounding completely natural."

Relatedness:
1: "Not at all related - The model's answer does not relate to the question, fails to follow the required format, or is outside the scope of possible answers."
2: "Somewhat related - The model's answer is related to the question to some extent and mostly follows the required format, staying generally within the scope of possible answers."
3: "Completely related - The model's answer is directly related to the question, follows the required format accurately, and fits within the scope of possible answers."

Correctness:
1: "Not at all correct - The answer is completely incorrect or irrelevant to the question posed."
2: "Somewhat correct - The answer is partially correct but includes some inaccuracies or incomplete information."
3: "Completely correct - The answer is fully correct, accurate, and provides a complete response to the question."

***
[END DATA]

Does the submission meet the criterion? First, write out in a step by step manner your reasoning about the criterion to be sure that your conclusion is correct. Avoid simply stating the correct answers at the outset. 
Your response must be RFC8259 compliant JSON following this schema:

{{"reasoning": str, "naturalness": int, "relatedness": int, "correctness": int}}
"""
\end{lstlisting}
\caption{The prompt for \GPTIV\ as evaluator. For \GPTIV\texttt{-no-gold} the gold answer is removed.}
\label{tab:gpt4prompt}
\end{figure*}

\end{document}